\documentclass[conference, 10 pt]{IEEEtran}  % Comment this line out if you need a4paper

\IEEEoverridecommandlockouts

\usepackage[font=small,belowskip=3pt,aboveskip=0pt]{caption}
\usepackage{titlesec}
\usepackage{float}
\usepackage{cite}
\usepackage{amsmath,amssymb,amsfonts}
\usepackage{algorithmic}
\usepackage{graphicx}
\usepackage{textcomp}
\usepackage[dvipsnames]{xcolor}

\usepackage{geometry}
\usepackage{booktabs}
\usepackage{ulem}
\def\BibTeX{{\rm B\kern-.05em{\sc i\kern-.025em b}\kern-.08em
    T\kern-.1667em\lower.7ex\hbox{E}\kern-.125emX}}

\usepackage{scalerel}
\usepackage{tikz}
\usetikzlibrary{svg.path}

\definecolor{orcidlogocol}{HTML}{A6CE39}
\tikzset{
  orcidlogo/.pic={
    \fill[orcidlogocol] svg{M256,128c0,70.7-57.3,128-128,128C57.3,256,0,198.7,0,128C0,57.3,57.3,0,128,0C198.7,0,256,57.3,256,128z};
    \fill[white] svg{M86.3,186.2H70.9V79.1h15.4v48.4V186.2z}
                 svg{M108.9,79.1h41.6c39.6,0,57,28.3,57,53.6c0,27.5-21.5,53.6-56.8,53.6h-41.8V79.1z M124.3,172.4h24.5c34.9,0,42.9-26.5,42.9-39.7c0-21.5-13.7-39.7-43.7-39.7h-23.7V172.4z}
                 svg{M88.7,56.8c0,5.5-4.5,10.1-10.1,10.1c-5.6,0-10.1-4.6-10.1-10.1c0-5.6,4.5-10.1,10.1-10.1C84.2,46.7,88.7,51.3,88.7,56.8z};
  }
}

\newcommand\orcidicon[1]{\href{https://orcid.org/#1}{\mbox{\scalerel*{
\begin{tikzpicture}[yscale=-1,transform shape]
\pic{orcidlogo};
\end{tikzpicture}
}{|}}}}

\usepackage{hyperref} %<--- Load after everything else

\usepackage{atbegshi}
\usepackage[absolute,overlay]{textpos}
\usepackage{rotating}
\usepackage[english]{babel}
\usepackage{blindtext}

\usepackage{framed}

\begin{document}

\title{A Cost-Effective and Climate-Resilient Air Pressure System for Rain Effect Reduction on Automated Vehicle Cameras}

\vspace{-0.5cm}

\author{Mohamed Sabry\orcidicon{0000-0002-9721-6291} \textit{Graduate Student Member, IEEE}, Joseba Gorospe\orcidicon{0000-0003-0334-5509} \textit{Member, IEEE} \\ and Cristina Olaverri-Monreal\orcidicon{0000-0002-5211-3598} \textit{Senior Member, IEEE}%

\vspace{-0.75cm}

% Johannes Kepler University
\thanks{Department Intelligent Transport Systems, Johannes Kepler University Linz, Altenberger Straße 69, 4040 Linz, Austria.
\texttt{
        \{mohamed.sabry, joseba.gorospe, cristina.olaverri-monreal\}@jku.at
    }
}
}

\maketitle

\vspace{-5pt}

\begin{abstract}

Recent advances in automated vehicles have focused on improving perception performance under adverse weather conditions; however, research on physical hardware solutions remains limited, despite their importance for perception critical applications such as vehicle platooning. Existing approaches, such as hydrophilic or hydrophobic lenses and sprays, provide only partial mitigation, while industrial protection systems imply high cost and they do not enable scalability for automotive deployment.

To address these limitations, this paper presents a cost-effective hardware solution for rainy conditions, designed to be compatible with multiple cameras simultaneously. 

Beyond its technical contribution, the proposed solution supports sustainability goals in transportation systems. By enabling compatibility with existing camera-based sensing platforms, the system extends the operational reliability of automated vehicles without requiring additional high-cost sensors or hardware replacements. This approach reduces resource consumption, supports modular upgrades, and promotes more cost-efficient deployment of automated vehicle technologies, particularly in challenging weather conditions where system failures would otherwise lead to inefficiencies and increased emissions. The proposed system was able to increase pedestrian detection accuracy of a Deep Learning model from 8.3\% to 41.6\%.

% The proposed system uses conventional materials and offers a scalable solution for automated vehicle platforms, effectively reducing the impact of adverse weather conditions.

\end{abstract}

\begin{IEEEkeywords}
Automated Vehicles, Adverse Weather Conditions, Perception, Sustainability
\end{IEEEkeywords}

%%%%%%%%%%%%%%%%%%%%%%%%%%%%%%%%%%%%%%
\begin{textblock*}{18.15cm}(1.55cm,26cm) % Adjust width and position as needed
\begin{minipage}{17.8cm}
     \vspace{0.1cm} % Vertical space within the minipage
     {\footnotesize\copyright 2026 IEEE. Personal use of this material is permitted. Permission from IEEE must be obtained for all other uses, in any current or future media, including reprinting/republishing this material for advertising or promotional purposes, creating new collective works, for resale or redistribution to servers or lists, or reuse of any copyrighted component of this work in other works. DOI: 10.1109/FISTS67319.2026.11421741}
\end{minipage}
\end{textblock*}
%%%%%%%%%%%%%%%%%%%%%%%%%%%%%%%%%%%%%%

% \vspace{-5pt}

\section{Introduction}

%Automated vehicles have seen significant advancements, offering substantial potential to enhance traffic safety, operational efficiency, and accessibility. Over the past decade, research and development in this domain have advanced rapidly, with a particular emphasis on refining perception modules that underpin autonomous functionalities. These modules incorporate multiple sensor modalities, including cameras, alongside algorithms capable of performing real-time perception tasks such as object detection and traffic signal recognition. However, the operational performance of such systems degrades substantially under adverse weather conditions, including rain, snow and fog. These conditions can impair sensor functionality, induce image distortion, and compromise data reliability, thereby introducing significant risks to both vehicle operation and passenger safety.

%A critical yet often overlooked challenge persists in the physical limitations of the sensors and cameras themselves. Adverse weather conditions can physically obstruct or degrade these components, significantly reducing the effectiveness of even advanced algorithms \cite{elmassik2022understanding}. Fig. \ref{fig:why_need_a_system} shows an example in which heavy rain directly affects the visibility of a camera located in an autonomous vehicle.

Automated vehicles have seen significant advancements over the past decade, offering substantial potential to enhance traffic safety, efficiency, and accessibility. These modules incorporate multiple sensor modalities, including cameras, to perform real-time perception tasks such as object detection and traffic signal recognition. However, their operational performance degrades under adverse weather conditions, including rain, snow, and fog, which can impair sensor functionality, induce image distortion, and compromise data reliability \cite{ZHANG2023146, elmassik2022understanding}.

This degradation is particularly critical in cooperative driving scenarios such as platooning \cite{zang2019rainplatoon}, where reliable and continuous detection of the lead vehicle is required to maintain safe and stable longitudinal control and prevent unsafe inter-vehicle spacing. %, thereby introducing risks to both vehicle operation and passenger safety. {\color{red} In the case of platooning, where a vehicle is connected with a lead vehicle in front, any distortion or inconsistency of tracking the lead vehicle can cause safety critical mishaps.} {\color{blue} TODO: SENTENCE for Platooning }

%{\color{blue} what do you think of the added paragraph  added before? }
% TODO: I'll add paragraph to link to platooning

From a sustainability perspective, reduced sensor reliability under adverse weather conditions can lead to higher maintenance demands through regularly applied solutions~\cite{pao2023method}, all of which contribute to greater material consumption which can have negative environmental impacts. Moreover, unreliable perception systems may force automated vehicles to disengage entirely, or cause accidents \cite{gruyer2021connected}, resulting in inefficient traffic flow, thus increased energy consumption, and higher emissions. Improving sensor robustness in challenging weather conditions is therefore not only a safety requirement but also a key enabler of sustainable and resource-efficient automated transportation systems.

% Figure \ref{fig:why_need_a_system} shows an example in which heavy rain directly affects the visibility of a camera located in an autonomous vehicle.

Despite this challenge, there remains a significant lack of physical solutions that can be seamlessly integrated into existing automated vehicle platforms to mitigate the impact of adversarial weather conditions. Current solutions often incur high cost, require repeated application due to degradation over time, or lack validation in realistic, dynamic driving scenarios. The absence of rigorous testing under environmental disruptions limits their practical applicability in real-world Intelligent Transportation System (ITS) deployments.

%There remains a deficiency in the development of physical solutions capable of seamless integration into existing automated systems to mitigate the effects of inclement weather. Existing solutions are either too expensive, requires repetitive application as its effect degrades with time. Furthermore, the aforementioned solutions were not tested in moving vehicles at 50+ Km/h to prove their effectiveness. 

To address this gap, in this paper, a physical Air Pressure System (APS) is presented with the capability of reducing the impact of rain on camera images. The proposed system adopts a centralized architecture, capable of supporting multiple sensors and allowing flexible swapping options for the main air distribution unit. Preliminary evaluations suggest that the APS significantly improves camera image clarity under adverse weather conditions, highlighting its potential as a  scalable and cost-efficient solution for climate-resilient automotive and ITS applications.

This paper is structured as follows: Section II reviews related work, Section III provides a detailed description of the proposed APS, Section IV presents the experimental setup followed by section V showing the results. Finally, Section VI summarizes the contribution of this work and outlines potential directions for future research.

%/////////////////////////////////////////////////

% \vspace{-3pt}

% \begin{figure}[!]
% 	\centering
% 	\includegraphics[width=1.0\columnwidth]{figures/Why_air.pdf}
% 	\caption{
% 		(A) illustrates an illumination invariant image to reduce the effects of varying weather conditions as presented in \cite{sabry2025shadow}. (B) shows the normal RGB image that is heavily influenced by rain even with a barrier to reduce the rain effect on the camera image. 
% 	} 
% 	\label{fig:why_need_a_system}
% \end{figure}

% \vspace{-5pt}

\section{Related Work}
\label{sec:RelatedWork} 
% TODO CONTINUE FROM HERE
% \vspace{-5pt}

Adverse weather conditions have a significant impact on camera-based perception systems, degrading image quality and reducing detection reliability in automated vehicles \cite{zang2019rainplatoon, elmassik2022understanding, YONEDA2019253}. While software-based approaches can partially mitigate these effects \cite{sabry2025shadow}, they remain limited when sensor performance is physically compromised. This section therefore reviews prior work on rain-effect mitigation, with a focus on industrial protection systems, consumer-grade liquid solutions, and specialized camera lenses.%Given the effects of adverse weather conditions on camera results, as mentioned in \cite{elmassik2022understanding}, as well as the partial improvements done through software modules, such as \cite{sabry2025shadow}, this section presents previous work on rain effect reduction, focusing on industrial approaches, consumer grade liquid solutions and specialized camera lenses. 
% TODO: I'll try to look for a reference or two extra

% \vspace{-3pt}
% \subsection{Camera Lenses}

Several water-repellent solutions for cameras have been proposed, including hydrophilic and hydrophobic surfaces designed to reduce water droplets forming on camera surfaces, as proposed in \cite{pao2022wind}. While these lenses can reduce the immediate effects of rain, they fail to provide comprehensive protection for continuous operation. In the aforementioned two papers, a simulation of a vehicle driving through rain confirmed the effect of rain on the camera image with both presented lenses.

An additional lens cleaning system is the Ultrasonic Lens Cleaning (ULC) system \cite{texasInstruments}. 
% \footnote{ \url{https://www.ti.com/tool/ULC1001-DRV290XEVM} }. 
This system is an evaluation module with a two-chip solution based on a piezo driver which vibrates to repel water droplets from the camera lens. The effectiveness of the system is hindered by requiring its own power supply, control boards and drivers, as well as limited to a single camera use. The price for each unit surpasses 100 euros. 

% \subsection{Temporary solutions}
A temporary solution for repelling rain from camera lenses is the use of water-repellent and anti-fog sprays \cite{rainX}. However, their utility is constrained by limited effectiveness and the necessity for frequent maintenance.

% \vspace{-5pt}

% \subsection{Industrial Solutions}
%In industrial use cases particularly in the film industry, high-end protective measures, such as heated enclosures and specialized coatings, are employed to safeguard cameras against harsh environmental conditions. However, these solutions are generally impractical for automated vehicles due to their high cost \footnote{ \url{https://www.berger-bros.com/accessories/movmax-hurricane-rain-deflector-pro/} }, complexity, and physical bulk. The implementation of such systems for individual vehicle cameras is not only economically infeasible but also incompatible with the demands of mass production and widespread deployment. Industrial solutions, while effective, lack scalability within the automotive sector, as their high cost and specialized requirements render them unsuitable for integration into consumer vehicles, where cost-efficiency and simplicity are essential.

%What about this:
In industrial applications, particularly within the film industry, high-end protective measures such as heated enclosures and specialized optical coatings are commonly employed to protect cameras operating under harsh environmental conditions. Although such solutions are effective when applied to individual cameras, they are generally incompatible due to their power requirements and high-cost (often reaching several thousand euros per unit \cite{rainDeflector}).
% \footnote{ \url{https://www.berger-bros.com/accessories/movmax-hurricane-rain-deflector-pro/} }).

% , as well as the complexity of their mechanical and thermal design, which demands precise design and maintenance. 
Consequently, while industrial-grade solutions demonstrate strong performance, they lack scalability for the automotive sector, where minimizing cost and adopting generalized, easily integrable solutions are required.

In response to these challenges, this study presents a cost-effective, weather-resilient solution designed for simple integration into existing automated vehicle systems. The proposed prototype system addresses the shortcomings of current solutions by being generic, cost-effective, and robust against a wide range of adverse weather conditions. 
% {\color{red}\sout{ A notable advantage of this system is its compatibility with multiple cameras simultaneously, enhancing its versatility and practicality for testing in the rain. The system employs affordable materials and design principles to establish a continuous air barrier that minimizes the impact of rain on the image quality of the camera. By prioritizing simplicity and cost-effectiveness, the system significantly enhances camera clarity in rainy conditions.}}

%The scalability of the solution facilitates its adoption across diverse vehicle models and manufacturers, thereby promoting broad implementation.
%Furthermore, the proposed system offers a viable alternative to expensive industrial solutions, rendering advanced weather protection accessible to a wider range of vehicles.

% \vspace{-3pt}

\section{The Proposed Air Pressure System} 
\label{sec:validatingSimulation}

% \vspace{-2pt}

\begin{figure*}
	\centering
	\includegraphics[width=0.85\textwidth]{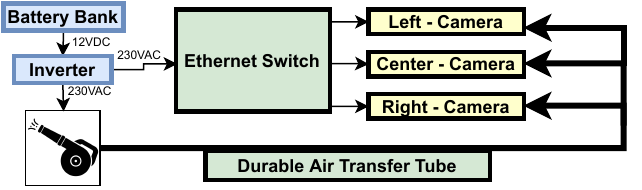}
	\caption{
		The figure illustrates the connectivity of the proposed Air Pressure System (APS) with the JKU-ITS research vehicle and the associated sensors to be cleaned. 
	}
	\label{fig:The_Air_pressure_system}
\end{figure*}

% \vspace{-2pt}

%The following section introduces the proposed APS which is plug and play and utilizes common materials that are widely available. The overall system architecture is depicted in Figure \ref{fig:System_Setup}. The APS is applied on the JKU-ITS \cite{certad2022jku} research vehicle, where the air pressure device is connected to the onboard power inverter with a 220V output. Weather resilient tubing is further connected to the device using quick connects, which enables quick disassembly and modifications. The main Tube is then connected to a series of standard Y connectors to split the air pressure among the available cameras. At each sensor lens, a custom flat air nozzle end was handmade to direct the air on the lens creating an air curtain to reduce the effect of water droplets on the lens. 

The following section presents a detailed overview of the proposed Air Pressure System (APS), a modular and user-friendly solution designed for integration with minimal setup requirements. The system leverages widely accessible, off-the-shelf materials, ensuring both cost-effectiveness and ease of implementation. A schematic representation of the overall system architecture is provided in Figure \ref{fig:The_Air_pressure_system}.

The APS was deployed on the JKU-ITS research vehicle \cite{certad2022jku}, where the core air pressure device is interfaced with the vehicle’s onboard power inverter, which supplies a standard 220V AC output. To facilitate efficient airflow distribution, weather-resilient tubing is connected to the device using quick-connect fittings, enabling rapid assembly, disassembly, and system modifications as needed. This design choice enhances operational flexibility and simplifies maintenance procedures. The air pressure device delivers a maximum air speed of 338 km/h and a maximum air flow of 12 m³/min, which is directly connected to a funnel connector to pass the air through to a 0.5 inch weather resilient tubing which then directs the air to the sensors.

The primary tubing is subsequently linked to a series of standard Y-connector junctions, which serve to divide and direct the pressurized airflow uniformly across the cameras mounted on the vehicle. At the terminal end of each branch, adjacent to the sensor lenses, a custom-fabricated nozzle was created by combining a 90° tube connector with a flat air exhaust, ensuring that the airflow is directed onto the camera lens at an obtuse angle, and therefore generating a controlled air curtain. This air curtain effectively disrupts and disperses water droplets upon contact, thereby mitigating their adverse effects on optical clarity and ensuring less-interrupted camera performance under extreme weather conditions. The modularity of the system, combined with its use of standardized components, allows for scalable deployment and adaptability to various sensor configurations.

\begin{figure}[!]
	\centering
	\includegraphics[width=0.8\columnwidth]{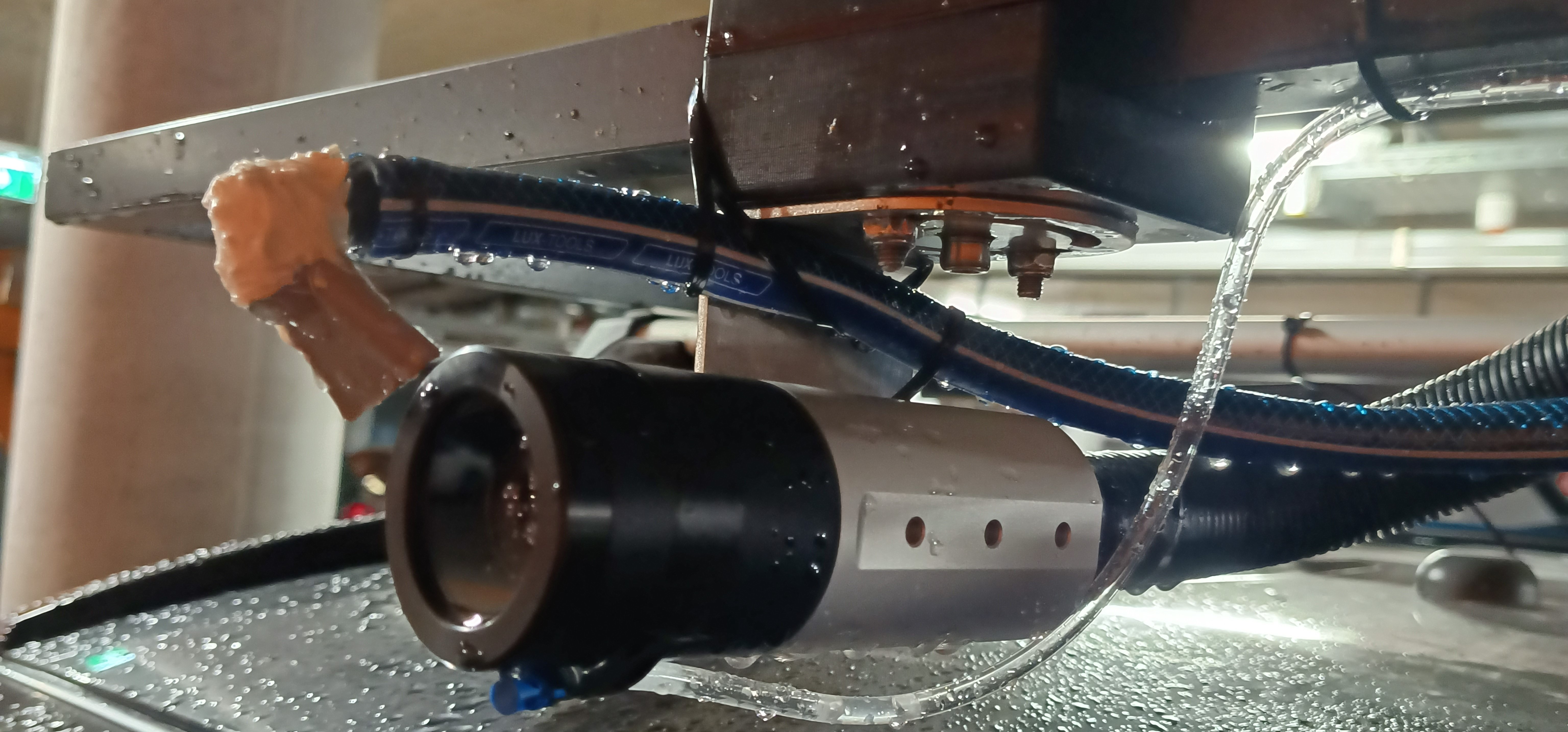}
	\caption{
		The figure illustrates the APS system mounted on the JKU ITS research vehicle. 
	} 
	\label{fig:The_camera_mount}
\end{figure}

Figure \ref{fig:The_camera_mount} shows the APS system tubing added to the center camera on the JKU-ITS research vehicle. The proposed APS has a total cost below 100 euros, including the main air pressure unit, the weather resilient tubes as well as the fixtures and fittings.

\vspace{-1pt}

% \vspace{-3pt}

%
%\begin{figure}[!h]
%	\centering
%	\includegraphics[width=1.0\columnwidth]{figures/Why_air.pdf}
%	\caption{
%		TODO NEED TO TAKE A GOOD IMAGE OF THE VEHICLE !!! AND SHOW THE AIR PRESSURE SYSTEM
%	} 
%	\label{fig:The system implemented on the vehicle}
%\end{figure}

% \section{ { \color{red} Experimental Setup \sout{ Experiments and Results } } }
% \vspace{-3pt}
% \vspace{-0.3cm}

\section{ Experimental Setup }
\label{sec:ExperimentalResults}

% \vspace{-0.15cm}
% {\color{red}\sout{and the results generated from the proposed fallback longitudinal controller.} }

% \subsection{The Simulation Scenario}

To evaluate the performance of the proposed APS, data was acquired using the center camera. Multiple recordings were conducted, both with and without the APS. Pedestrian detection was applied to images using the YOLOv4-tiny Deep Learning (DL) Network \cite{bochkovskiy2020yolov4} for its balance between computational efficiency and robustness.

Two evaluation methods were employed to assess if APS can reduce the impact of the rain to, in this case, a pedestrian DL detection model: 1) Frame-based qualitative evaluation, where a test is considered successful if a pedestrian was detected across three consecutive frames, and 2) time-based quantitative evaluation, were a 10 second interval in which the pedestrian was within the camera's field of view, the percentage of frames where it was correctly detected was calculated. The aim of the qualitative evaluation method is to visually assess if the vehicle with APS is able to continuously detect and track targets, while the quantitative approach provides an overall performance metric. These evaluation methods were chosen as in adverse weather conditions, one of the critical issues is continuous tracking without dropping frames. Especially in safety critical scenarios such as platooning, with vehicles maintaining short inter-vehicle distances at speeds exceeding 50 km/h, real-time tracking of the lead vehicle is essential for stable and safe operation.
%{\color{red} The aforemetioned testing case was conducted, as the proposed APS will be integrated in a platooning case, where the JKU-ITS research vehicle will be autonomously following a lead vehicle in rainy weather conditions. In real-time operation, if the lead vehicle detection becomes inconsistent the stability of the platoon directly gets affected, which in turn increases the safety risk of the operation. In the case where the rain droplets cover vital areas on the camera lens without being removed fast enough, such as the location where the lead vehicle is seen by the DL model, the autonomous system will lose the camera-based vehicle detection. This results in disabling the long range and fusion detection module, in turn, significantly reducing the safety of operation due to having no redundancy when tests are conducted at 50+ Km/h with inter-vehicular distances less than 11 meters.}

% resulting a the images  mimicing a grayscale output instead of color.}

\section{ Results }

% \vspace{-5pt}
\begin{figure*}
	\centering
	\includegraphics[width=0.9\textwidth]{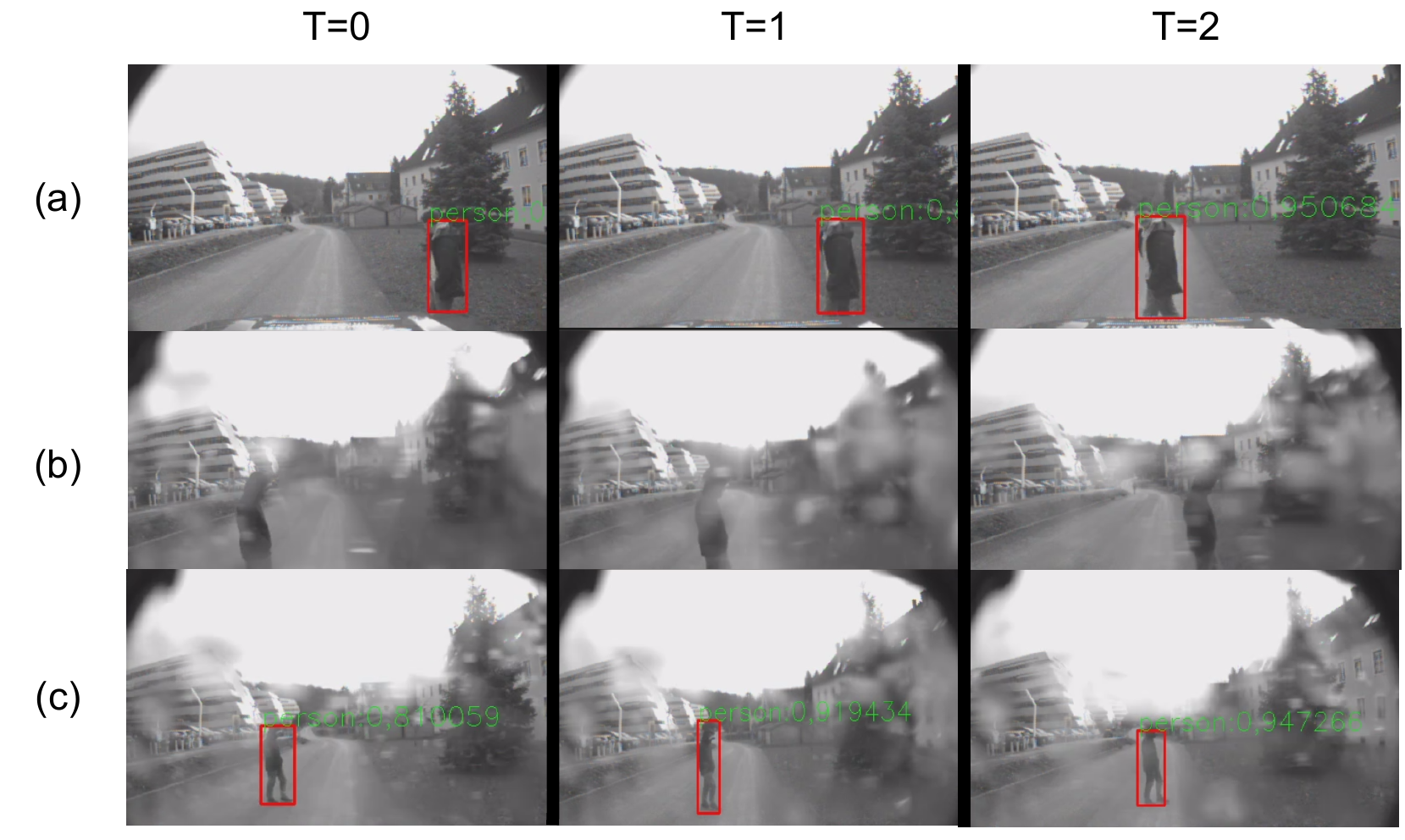}
	\caption{
		The figure illustrates the results from using the proposed APS system. The figure shows there consecutive frames taken from a Basler camera with Yolov4-tiny used as a Deep learning model to test the effect of the system. (a) shows the results from a clean camera lens. (b) shows the results in rain without the proposed system. (c) shows the results while raining and using the proposed APS.
	}
	\label{fig:The_RESULTS}
    \vspace{-5pt}
\end{figure*}

Figure \ref{fig:The_RESULTS} presents the qualitative results of the experiments conducted in December during rainy weather with dense clouds, showing both image quality and pedestrian detection outcomes. A successful test in this case is denoted by correct pedestrian detection across 3 consecutive frames. Subfigure (a) shows the baseline results on a clear lens without the APS. Subfigure (b) illustrates the impact of rainy weather, which noticeably degrades the quality of the image, as well as the performance of the DL model. Subfigure (c) shows  the results while using the APS during rain. It can be noted that the APS significantly improved detection performance of the DL under adverse weather conditions.

%the DL pedestrian detection results are shown in 3-consecutive frames on a clear lens without the system (Subfigure (a) ). Subfigure (b) shows the results in rainy weather. It can be noted that the rain significantly affected the DL results. In Subfigure (c) and Subfigure (d) both show the results while using the APS system while it is raining (d) and while the rain gradually decreased in density. It can be noted that the APS system significantly improved the results of the DL model.

Table \ref{tab:frame_det} presents quantitative results comparing the baseline detection percentage, rainy conditions without APS and rainy conditions with APS. The results are derived from the same experiment depicted in Fig. \ref{fig:The_RESULTS}, with each condition recorded continuously for 100 seconds. The data were divided into 10-second intervals, and the table reports the average pedestrian detection rate across all intervals for each scenario. The APS system substantially improved the performance of the perception system, achieving a 41.6\% detection rate across a 10 second interval in rainy weather conditions, compared to only 8.3\% without the proposed system.

\vspace{-5pt}

\renewcommand{\tabcolsep}{3pt}
\begin{table}[h]
\centering
\caption{ Quantitative results across 10 seconds}
\label{tab:frame_det}
\begin{tabular}{|c|c|}

\hline
\textbf{Weather Condition} & \textbf{Detection \% } \\ \hline
\begin{tabular}[c]{@{}c@{}} Baseline No Rain \end{tabular}  & 100  \\ \hline
\begin{tabular}[c]{@{}c@{}}Rain Without APS \end{tabular} & 8.3       \\ \hline
\begin{tabular}[c]{@{}c@{}}Rain With APS \end{tabular} & 41.6         \\ \hline
\end{tabular}
\end{table}

% \vspace{-2pt}

\section{Conclusion and Future Work}
\label{sec:Conclusions}

The proposed APS effectively reduces the impact of rain on camera images,  demonstrating its applicability for cameras mounted outside a vehicle and exposed to adverse weather conditions. The APS, with a total cost below 100 euros, is cost-effective, easy to install, and capable of supporting multiple sensors simultaneously.

Future research will focus on improving the system's components to simplify disassembly, and extending its functionality to other sensors, such as LiDAR, enhancing the versatility and scalability of the APS for broader ITS systems. Further analysis will aim to maximize airflow transmission efficiency, and more exhaustive testing will be conducted under a variety of weather conditions.

\vspace{-3pt}

\section*{Acknowledgment}

% \vspace{-3pt}

This work was funded by the Austrian Research Promotion Agency (FFG) project pDrive number: 12451001.

% \vspace{-0.5cm}

\bibliographystyle{IEEEtran}
\bibliography{IEEE-conference-template-062824}

@article{gruyer2021connected,
  title={Are connected and automated vehicles the silver bullet for future transportation challenges? Benefits and weaknesses on safety, consumption, and traffic congestion},
  author={Gruyer, Dominique and Orfila, Olivier and Glaser, S{\'e}bastien and Hedhli, Abdelmename and Hauti{\`e}re, Nicolas and Rakotonirainy, Andry},
  journal={Frontiers in sustainable cities},
  volume={2},
  pages={607054},
  year={2021},
  publisher={Frontiers Media SA}
}

@misc{rainDeflector,
  title = {MOVMAX Hurricane Rain Deflector Pro},
  note = {Accessed December 19, 2025.  \url{https://www.berger-bros.com/accessories/movmax-hurricane-rain-deflector-pro/}}
}

@misc{rainX,
  title = {Rain-X® Original Glass Water Repellent Aerosol},
  note = {Accessed December 19, 2025.  \url{https://www.rainx.com/product/rain-x-original-glass-water-repellent-aerosol/}}
}

@misc{texasInstruments,
  title = {Texas Instruments Ultrasonic Lens},
  note = {Accessed December 19, 2025.  \url{https://www.ti.com/tool/ULC1001-DRV290XEVM}}
}

@article{bochkovskiy2020yolov4,
  title={Yolov4: Optimal speed and accuracy of object detection},
  author={Bochkovskiy, Alexey and Wang, Chien-Yao and Liao, Hong-Yuan Mark},
  journal={arXiv preprint arXiv:2004.10934},
  year={2020}
}

@inproceedings{certad2022jku,
  title={J{KU}-{ITS} automobile for research on autonomous vehicles},
  author={Certad, Novel and Morales-Alvarez, Walter and Novotny, Georg and Olaverri-Monreal, Cristina},
  booktitle={International Conference on Computer Aided Systems Theory},
  pages={329--336},
  year={2022},
  organization={Springer}
}

@article{sabry2025shadow,
  title={Shadow Erosion and Nighttime Adaptability for Camera-Based Automated Driving Applications},
  author={Sabry, Mohamed and Schroeder, Gregory and Varughese, Joshua and Olaverri-Monreal, Cristina},
  journal={arXiv preprint arXiv:2504.08551},
  year={2025}
}

@article{pao2023method,
  title={A method of evaluating ADAS camera performance in rain--case studies with hydrophilic and hydrophobic lenses},
  author={Pao, Wing Yi and Li, Long and Agelin-Chaab, Martin},
  journal={Progress in Canadian Mechanical Engineering},
  volume={6},
  pages={354},
  year={2023}
}

@article{pao2022wind,
  title={Wind-driven rain effects on automotive camera and LiDAR performances},
  author={Pao, Wing Yi and Li, Long and Agelin-Chaab, Martin},
  year={2022}
}

@inproceedings{elmassik2022understanding,
  title={Understanding the Scene: Identifying the Proper Sensor Mix in Different Weather Conditions.},
  author={Elmassik, Ziad and Sabry, Mohamed and El Mougy, Amr},
  booktitle={ICAART (3)},
  pages={785--792},
  year={2022}
}

@ARTICLE{zang2019rainplatoon,

  author={Zang, Shizhe and Ding, Ming and Smith, David and Tyler, Paul and Rakotoarivelo, Thierry and Kaafar, Mohamed Ali},

  journal={IEEE Vehicular Technology Magazine}, 

  title={The Impact of Adverse Weather Conditions on Autonomous Vehicles: How Rain, Snow, Fog, and Hail Affect the Performance of a Self-Driving Car}, 

  year={2019},

  volume={14},

  number={2},

  pages={103-111},

  doi={10.1109/MVT.2019.2892497}}

@article{YONEDA2019253,
title = {Automated driving recognition technologies for adverse weather conditions},
journal = {IATSS Research},
volume = {43},
number = {4},
pages = {253-262},
year = {2019},
issn = {0386-1112},
doi = {https://doi.org/10.1016/j.iatssr.2019.11.005},
url = {https://www.sciencedirect.com/science/article/pii/S0386111219301463},
author = {Keisuke Yoneda and Naoki Suganuma and Ryo Yanase and Mohammad Aldibaja}
}

@article{ZHANG2023146,
title = {Perception and sensing for autonomous vehicles under adverse weather conditions: A survey},
journal = {ISPRS Journal of Photogrammetry and Remote Sensing},
volume = {196},
pages = {146-177},
year = {2023},
issn = {0924-2716},
doi = {https://doi.org/10.1016/j.isprsjprs.2022.12.021},
url = {https://www.sciencedirect.com/science/article/pii/S0924271622003367},
author = {Yuxiao Zhang and Alexander Carballo and Hanting Yang and Kazuya Takeda}
}

\end{document}